\documentclass[runningheads]{llncs}
\usepackage{graphicx}
\usepackage{comment}
\usepackage{amsmath,amssymb} 
\usepackage{color}
\usepackage{caption}
\usepackage{subfig}
\usepackage{enumerate}
\usepackage{authblk}
\usepackage{tikz}
\usepackage{booktabs}

\usepackage{dsfont}
\usepackage[hidelinks]{hyperref}

\begin{document}
\pagestyle{headings}
\mainmatter
\def\ECCVSubNumber{6490}  
\title{Sketch-Guided Object Localization \\in Natural Images}
\author{Aditay Tripathi$^{1}$,
Rajath R Dani$^{1}$,
Anand Mishra$^{2}$ and 
Anirban Chakraborty$^{1}$}
\institute{$^{1}$Indian Institute of Science, Bengaluru~~~$^{2}$Indian Institute of Technology, Jodhpur  \\
\href{http://visual-computing.in/sketch-guided-object-localization/}{http://visual-computing.in/sketch-guided-object-localization/}}


\titlerunning{Sketch-Guided Object Localization in Natural Images}
\authorrunning{Tripathi et al.}
\maketitle
\begin{abstract}
We introduce the novel problem of localizing all the instances of an object (seen or unseen during training) in a natural image via sketch query. We refer to this problem as sketch-guided object localization. This problem is distinctively different from the traditional sketch-based image retrieval task where the gallery set often contains images with only one object. The sketch-guided object localization proves to be more challenging when we consider the following: (i) the sketches used as queries are abstract representations with little  information on the shape and salient attributes of the object, (ii) the sketches have significant variability as they are hand-drawn by a diverse set of untrained human subjects, and (iii) there exists a domain gap between sketch queries and target natural images as these are sampled from very different data distributions. To address the problem of sketch-guided object localization, we propose a novel \emph{cross-modal attention} scheme that guides the  region proposal network (RPN) to generate object proposals relevant to the sketch query. These object proposals are later scored against the query to obtain final localization. Our method is effective with as little as a single sketch query. Moreover, it also generalizes well to object categories not seen during training and is effective in localizing multiple object instances present in the image. Furthermore, we extend our framework to a multi-query setting using novel feature fusion and attention fusion strategies introduced in this paper. The localization performance is evaluated on publicly available object detection benchmarks, viz. MS-COCO and PASCAL-VOC, with sketch queries obtained from `Quick, Draw!'. The proposed method significantly outperforms related baselines on both single-query and multi-query localization tasks.
\end{abstract}

\begin{figure}[!h]
    \centering
    \includegraphics[width = 12 cm]{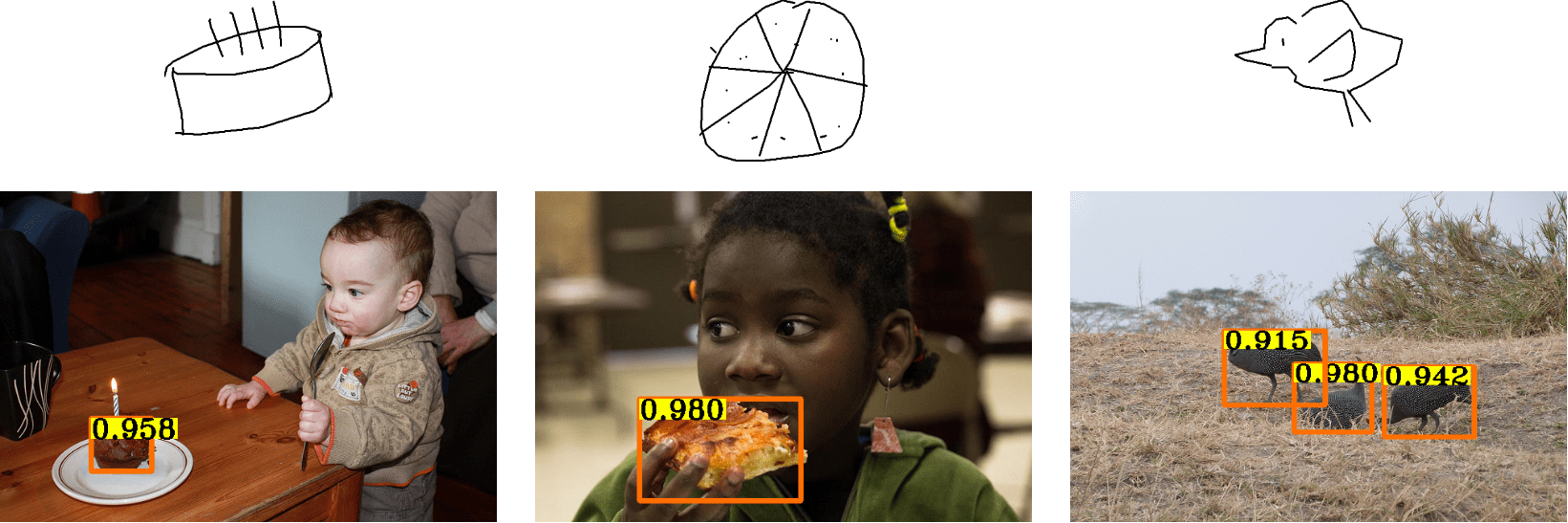}
    \caption{\label{fig:goal} \textbf{Sketch-guided object localization.} Can we localize a cake, a pizza and birds in these images by just drawing their sketches? In this paper, we introduce the problem of sketch-guided object localization, and propose a novel \emph{cross-modal attention} scheme to address this problem. \textbf{[Best viewed in color]}.}
\end{figure}
\section{Introduction}
\label{sec: intro}
Localizing objects in a scene via an image query has been a long-sought pursuit in computer vision literature. The seminal paper Video Google~\cite{sivic2003video} addresses object localization in videos using a text retrieval approach. More recent works on this problem focus on localizing seen as well as unseen object categories with as little as one image query during test time~\cite{hsieh2019one,li2018high}. There are several applications associated with object localization in an image via image query - examples include object tracking~\cite{li2018high} and content-based image retrieval~\cite{sharif2014cnn,wan2014deep}. However, the image of an object may not always be preferred as a query due to many practical reasons. These include (i) copyright issues, (ii) privacy concerns, or (iii) significant overhead to collect and annotate images of a rapidly expanding object category set, e.g., in industrial applications where images of each part of the equipment may not be available during the training. Further, it is also not practical to assume that the object names are always available for localization. In such situations, where stronger visual or semantic cues are unavailable, it would be worth exploring if a hand-drawn sketch of an object can be utilized for localizing the object in a natural image. In other words, given hand-drawn sketches of objects (for example, cake, pizza, and bird as shown in the top row of Fig.~\ref{fig:goal}) can we design a framework that learns to localize all instances of these objects present in natural images? In this work, we explore this problem for the first time in the literature and propose a solution to it. 

Despite the problem of sketch-guided object localization has a relation to the well-studied sketch-based image retrieval (SBIR) problem~\cite{eitz2010sketch,hu2013performance,liu2017deep,sangkloy2016sketchy,shen2018zero,song2017deep,yu2016sketch,zhang2018generative}, these are distinctively different in terms of objectives. It is important to note that SBIR aims to retrieve images from a gallery of localized objects for a given sketch query. Contrary to this, our objective is to precisely localize all instances of an object of interest in a natural scene in presence of many other distracting objects. Furthermore, sketch-guided object localization poses several additional challenges compared to image query-guided object localization, notably - (i) the sketches used as queries are abstract drawings with little information on shape and salient attributes of the object, (ii) the sketches have significant variability as they are drawn by a diverse set of untrained human subjects, and  (iii) a large domain gap exists between the sketch queries and the target gallery images.

\noindent\textbf{Plausible baseline approaches vs. proposed solution:}
Despite its practical utility, sketch-guided object localization has never been explored in the literature. In this work, we consider a state-of-the-art image query-guided localization method~\cite{hsieh2019one} and a modified Faster R-CNN~\cite{ren2015faster} as probable baseline approaches. We empirically show that these methods are insufficient due to their sensitivity to the domain gap present between sketches and natural images, and their ineffective attention mechanisms in this context. To mitigate these issues, we develop a \emph{cross-modal attention} scheme to guide proposal generation, which is a crucial step in the localization framework. The proposed \emph{cross-modal attention} scheme generates a spatial compatibility matrix by comparing the global sketch representation with the local image representations obtained from each location of the image feature map. This spatial compatibility matrix creates an attention feature map that is fused with the original feature map of the image. The result of this fusion is subsequently fed to a region proposal network (RPN) to generate proposals relevant to the sketch query. Finally, the proposals are pooled and compared against the query to precisely localize the object of interest.

Our proposed model, by virtue of the aforementioned \emph{cross-modal attention}, embeds query information in the image feature representation before generating region proposals. Therefore, it is inherently able to generate relevant object proposals even for object categories unseen during the training time. This enables the proposed method to achieve superior performance to baseline methods for unseen categories while achieving reasonably high performance for seen categories as well. We also explore the possibility of using multiple sketch queries towards the localization task. To generalize our proposed model to the multi-query setting, we propose novel feature and attention pooling mechanisms. 

\noindent\textbf{Contributions:} In summary, we make the following contributions:\\
\noindent(i) We introduce an important but unexplored problem -- \emph{sketch-guided object localization in natural images}. This novel problem is well motivated from scenarios where a query image corresponding to the object of interest or the object category names are not available, but a sketch can be hand-drawn to provide a minimal visual description of the object.

\noindent(ii) We propose a novel \emph{cross-modal attention} scheme to guide the region proposal network to generate object proposals relevant to the sketch query. Our method is effective with as little as one sketch used as a query. Moreover, it generalizes well to unseen object categories with $\approx 3\%$ improvement over a one-shot detection method used as baseline~\cite{hsieh2019one} and is also effective in localizing multiple object instances present in the image.

\noindent(iii) To support multiple sketch-query based object localization, we propose feature and attention pooling in our architecture and demonstrate promising performance.
    
\noindent(iv) We have performed extensive experimentation and ablation studies on two public benchmarks. We firmly believe that our work will open-up novel future research directions under sketch-guided computer vision tasks.

\section{Related work}
\label{sec: relwork}
\noindent\textbf{Sketch-based Image Retrieval:} 
Given a sketch query, sketch-based image retrieval (SBIR) aims to retrieve from a gallery of images containing a single object. Methods in SBIR can be grouped into two broad categories- (a) \textit{classical methods} which use hand-crafted features, for example, SIFT or gradient-field histogram of gradients along with a bag-of-words representation of sketches as shown in~\cite{eitz2010sketch,hu2013performance},  (b) \textit{deep methods} which utilize cross-modal deep learning techniques by incorporating ranking loss such as the contrastive loss~\cite{sangkloy2016sketchy} or the triplet loss~\cite{yu2016sketch} to learn a ranking function between candidate images and sketch queries which are then subsequently used to score the candidate images, and the top-scoring image is retrieved. In~\cite{song2017deep}, researchers have utilized an attention model to solve fine-grained SBIR, and have introduced HOLEF loss to bridge the domain gap between the sketches and images. Since the time complexity of such methods for large-scale SBIR problem is significantly high, researchers have proposed hashing models, such as~\cite{liu2017deep,shen2018zero,xu2018sketchmate,zhang2018generative}, which significantly reduces the retrieval time. Contrary to these works, our goal is to use a sketch query to \emph{localize} all instances of an object in an image that usually contains many objects of different categories.\\
\newline
\noindent \textbf{Object Detection:}
Object detection involves localizing and classifying an object in a given image. State-of-the-art object localization methods can be primarily grouped into two categories: \textit{proposal-free}~\cite{kong2019foveabox,lin2017focal,liu2016ssd,redmon2016you,redmon2018yolov3,sermanet2013overfeat} and \textit{proposal-based}~\cite{cai2018cascade,girshick2015fast,girshick2014rich,he2017mask,he2015spatial,ren2015faster}. \textit{Proposal-free} methods are single-stage detectors, and are faster during inference. However, they often fall short in performance as compared to \textit{proposal-based} methods. 
\textit{Proposal-based} methods first generate object proposals and then refine them by classifying each of them into one of the object categories. In~\cite{girshick2014rich}, one such method, selective search~\cite{uijlings2013selective} was utilized to generate proposals in the first stage, and then use these generated proposals to classify the object. However, the two-stages in this model were trained independently. Faster R-CNN~\cite{ren2015faster} introduced a region proposal network (RPN) that made the detection pipeline end-to-end trainable. Note that, both these types of object detectors are query-free. More recently, Hsieh et al.~\cite{hsieh2019one} have introduced one-shot object detection via an image query, where the goal is to localize all the instances of an unseen object in the target image via an image query of the same object. Unlike their work, where query and target images are from the same distribution, our queries, i.e., hand-drawn sketches, are from a significantly different domain as compared to that of the target images.\\
\newline
\noindent\textbf{Attention in Deep Neural Networks:} 
The attention model in deep neural networks helps the salient features of the input image to become more critical when required. In~\cite{choe2019attention,teh2016attention}, the authors proposed attention networks to generate attention scores on each of the object proposals, and an attention-based dropout layer for weakly supervised object localization. In~\cite{caicedo2015active} a dynamic attention-action strategy using deep reinforcement learning was proposed to localize objects in an image. Hsieh et al.~\cite{hsieh2019one} adapted the self-attention mechanism proposed in~\cite{wang2018non} and applied it to image query-guided object detection. In this method, each pixel in an image was represented as a weighted combination of the pixels in the query image. The weights depend on the similarity between each image pixel and query pixel pairs for all pixels in the query. Conversely, our proposed \emph{cross-modal attention} computes the spatial compatibility between global query representation and localized representations of image regions. This enables it to mitigate the domain misalignment problem prevalent in sketch-guided object localization. Yan et al.~\cite{yan2019meta} used class-specific attentive vectors inferred from images of objects in a meta-set for applying channel-wise soft attention to feature maps of the proposals. However, the channel-wise soft attention may not be trivially utilized in our case due to the domain gap between sketch query and target natural images. \\
\newline
\noindent\textbf{Sketch Representation:} 
Besides the traditionally used convolutional neural networks~\cite{yu2017sketch}, there have been some recent advancements towards robust representation learning for sketches, e.g., SketchRNN~\cite{ha2017neural}, transformer-based architecture~\cite{xu2019multi}. Although we have chosen to utilize CNN (ResNet)-based feature encoders in this work, our proposed localization framework can seamlessly integrate more advanced sketch representation learning methods in the feature extraction modules.

\section{Proposed Methodology}
\label{sec:model}
In this section, we first formally introduce the novel problem of sketch-guided object localization in natural images. Then, we present a solution framework built around our novel \textit{cross-modal attention} scheme and discuss its utility for one-shot object detection using a sketch query. Further, we extend our proposed scheme to a multi-query setting, i.e., \emph{multiple sketch query-based object localization} task by introducing principled fusion strategies. 

\subsection{Problem Formulation}
Let $I = \{I_{train}, I_{test}\}$ be a set of all natural images, each containing variable number of object instances and categories, in a dataset ${{\cal D}_I}$. Here $I_{train}$ and $I_{test}$ are sets of train and test images respectively. Like any other machine learning task these two sets are mutually exclusive and only $I_{train}$ is available during training time.  Further, let $S=\{S_{train}, S_{test}\}$ be a set of all sketches, each containing one object, and $C = \{C_{train}, C_{test}\}$ be a set of all object categories. During training time, each training sample contains an image $i \in I_{train}$, a query sketch  $s_c \in S_{train}$, where $c \in C_{train}$, and all the bounding boxes corresponding to object category $c$ in the image $i$. During test time, given an image
$i^{'} \in I_{test}$ and sketch query $s_c \in S_{test}$, where $c \in C_{test}$, the problem is to localize all the instances of the object category $c$ in the image $i^{'}$. Note that we show experimental results in cases where $C_{train} = C_{test}$ i.e., categories in $C_{test}$ are seen during training time (common train-test categories), as well as $C_{train} \cap C_{test} = \phi$, i.e., categories in $C_{test}$ are not seen during training (disjoint train-test categories). 

\begin{figure}[!t]
    \centering
    \includegraphics[width = 12 cm]{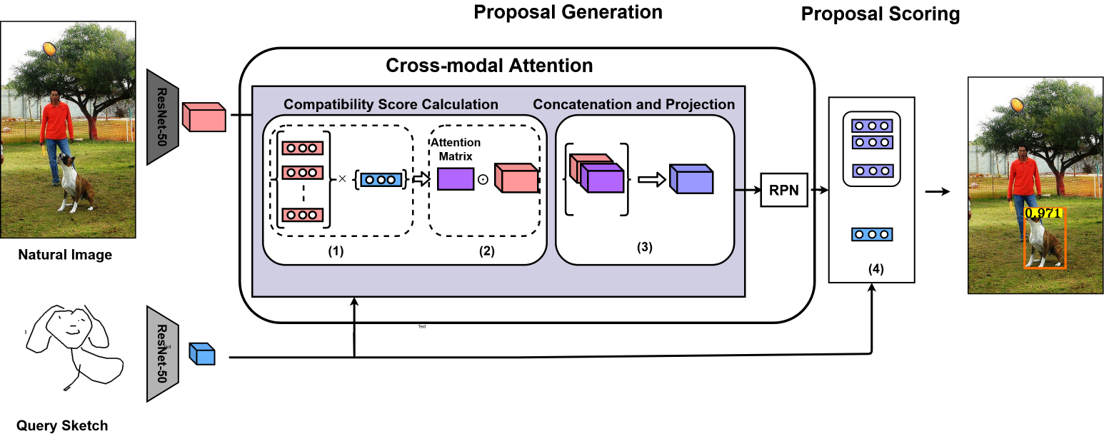}
    \caption{\label{fig:approach} Given an image and a query sketch, our end-to-end trainable sketch-guided localization framework works in the following two stages: (i) \textbf{query-guided proposal generation:} in this stage, feature vectors corresponding to different regions in the image feature map (shown using pink) are scored with the global sketch representation (shown using blue) to identify the compatibility (Block-1). Then, these compatibility scores (shown using violet) are multiplied with image feature maps (shown using pink) to get attention feature (Block-2). Further, these attention feature maps are concatenated with the original feature maps and projected to lower-dimensional space which is then passed through region proposal network to generate relevant object proposals (Block-3), (ii) \textbf{proposal scoring:} the pooled object proposals (shown using indigo) are scored with sketch feature vector (shown using blue) to generate localization for the object of interest (Block-4). \textbf{[Best viewed in color].}}
\end{figure}

The proposed sketch-guided object localization is an end-to-end trainable framework which works in the following two stages: (i) query-guided object proposal generation (Section~\ref{sec:attention}), (ii) proposal scoring (Section~\ref{sec:scoring}). Fig.~\ref{fig:approach} illustrates the overall architecture of the proposed framework.

\subsection{Cross-modal Attention for Query-guided Object Proposal Generation}
\label{sec:attention}
A popular object detection framework, viz. Faster R-CNN~\cite{ren2015faster} uses a region proposal network (RPN) module to generate object proposals. We can use the same module for generating object proposals in our task. However, the RPN is not designed to utilize any information on object appearance available in the query images. Hence, the object proposals that are relevant to the sketch query may not even be generated, especially when the object of interest is of low resolution, occluded or hidden among objects that are better represented in the input images. Therefore, using a RPN in its vanilla form may not be useful in our pipeline. To solve the aforementioned problem, we propose \emph{cross-modal attention} to incorporate the sketch information in the RPN to guide the proposal generation. Regions of interest (ROIs) are pooled from these region proposals using a strategy similar to the Faster R-CNN, and a scoring function $\Theta$ is learned between these ROIs and sketch queries.

We will now describe our novel \emph{cross-modal attention} scheme that we have introduced to generate object proposals relevant to an input query sketch. The attention module is trained to produce a weight map that provides high scores to the areas in an image that are similar to the given query sketch.

A sketch $s_c \in S$ of an object category $c \in C$ is used to query an image $i\in I$. To generate the feature representation of sketches and images, we use ResNet-50 models pretrained on Imagenet~\cite{deng2009imagenet} and Quick-Draw~\cite{jongejan2016quick} datasets for images and sketches respectively as a backbone. Suppose $\phi_I$ and $\phi_S$ represent these backbone models, then image and sketch feature maps are computed as: 

\begin{equation}
    i^{\phi_I} = \phi_I(i)\quad \text{and}\quad s^{\phi_S}_c = \phi_S(s_c),
\end{equation}
where, $i^{\phi_I} \in \mathbb{R}^{w\times h \times d}$ and $s^{\phi_S}_c \in \mathbb{R}^{w'\times h' \times d}$ are the extracted image and sketch feature maps respectively. From these feature maps, the compatibility score is learned between the sketch and the image feature maps by first applying non-linear transformations as below:
\begin{equation}
    i^{\psi_I} = \psi_I(i^{\phi_I})\quad \text{and} \quad s_{c}^{\psi_S} = \psi_S(s_{c}^{\phi_S}).
\end{equation}

A set of local feature vectors is formed by obtaining one vector at each location $(m,n)$ in the image feature map $i^{\psi_I}$, where $m\in \{1, 2, \dots, w\}$ and $n\in \{1,2,\dots, h\}$. Each vector represents a spatial region on the target image and the set gives us spatial distribution of the features. Subsequently, this is compared against a global representation of the sketch features. For image feature map i.e. $i^{\psi_I} \in \mathbb{R}^{w\times h \times d}$, the extracted set of feature vectors is represented as $L^i= \{\mathbf{L}_1^i, \mathbf{L}_2^i,...,\mathbf{L}_{w \times h}^i \}$ where $\mathbf{L}_j^i \in \mathbb{R}^{1\times 1\times d}$ $\forall$ $j\in \{1,2,\cdots w\times h\}$.

In the case of sketches, a global representation of sketch feature maps is obtained via the global max pool (${\cal GMP}$) operation, i.e.,

\begin{equation}
    \mathbf{L}_g^{s_c} = {\cal GMP} (s_c^{\psi_S}),
\end{equation}
where, $\mathbf{L}_g^{s_c} \in \mathbb{R}^{1\times 1 \times d}$.
A spatial compatibility score between $\mathbf{L}_j^i \in L^i$ and $\mathbf{L}_g^{s_c}$, is evaluated as follows:
\begin{equation}
\label{eq:constant}
    \lambda(\mathbf{L}^i_j, \mathbf{L}_g^{s_c}) = \frac{\left(\mathbf{L}_j^i \right)^T \left(\mathbf{L}_g^{s_c} \right)}{\cal K} \,\,\,\,\, ,
\end{equation}
where $\cal K$ is a constant. For simplicity of notation, we will refer to the left hand side of eq.(\ref{eq:constant}) as $\lambda_{jg}$ from here onwards.

Note that these compatibility scores are generated as a spatial map, which can be understood as a weight map representing attention weights. Therefore, in order to obtain attended feature maps, we perform element-wise multiplication of these compatibility scores and the original image feature map at each spatial location, i.e.,
\begin{equation}
    i^{a_I}_j = i_{j}^{\phi_I}\odot \lambda_{jg}, 
\forall j \in \{1,2,\cdots,w \times h\}.
\end{equation}
This attention feature map aims to capture information about the location of objects in an image that shares high compatibility score with the sketch query. Therefore, to incorporate this information, attention feature maps are concatenated along the depth with the original feature maps, i.e., $i^{\phi_I}_f = [(i^{a_I})^T;(i^{\phi_I})^T]^T$, where $i_f^{\phi_I}\in \mathbb{R}^{w\times h \times 2d}$. These concatenated feature maps are projected to a lower dimensional space to obtain the final feature maps, which are subsequently passed through the RPN to generate object proposals relevant to the sketch query. 

\subsection{Proposal Scoring}
\label{sec:scoring}
Once a small set of proposals, represented as $R_i$ for $i\in I$, are pooled from all query-guided region proposals generated by RPN, a scoring function $\Theta$ is learned to rank these proposals with respect to the sketch query. To this end, during the training phase, each of these region proposals are labeled as foreground (1) if it has $\geq 0.5$ intersection over union (IoU) with any of the ground-truth bounding boxes and the object in the ground-truth bounding box is the same class as sketch query, and background (0) otherwise. Then, we minimize a margin rank loss between the generated object proposals and the sketch query such that object proposals similar to the query sketch are ranked higher. 

To learn the scoring function $\Theta$, firstly, we generate feature vectors for the sketch query and the object proposals by taking a global mean pool operation on the sketch feature maps and object proposal feature maps, respectively. Each of the proposal feature vectors is concatenated with the sketch feature vector. These concatenated feature vectors are passed through a scoring function and the foreground probabilities of the proposals in context with the sketch query are predicted. Let $a_k$ be the predicted foreground probability for proposal $r_k \in R_i$, and it is given by:

\begin{equation}
\label{eq:concat}
    a_k = \Theta([g_m(r_k^{\phi'_I})^T;g_m({s_c^{\phi'_S}})^T]^T),
\end{equation}
where, $r_k^{\phi'_I}$ is the feature map for $r_k \in R_i$ generated using standard Faster R-CNN protocols and $g_m: \mathbb{R}^{W\times H\times D} \rightarrow \mathbb{R}^{1\times 1\times D}$ is the global mean pool operation. Now, towards training the scoring function, a label $y_k=1$ is assigned to $r_k$ if it is part of any foreground bounding box and $y_k=0$, otherwise. Motivated from~\cite{hsieh2019one} the loss function used in training is defined as:
\begin{align}
\label{eq:ce}
    L(R_i, s_c) = &\sum_{k}\left \{y_k  \max(m^+-a_k,0) + (1-y_k) \max(a_k-m^-,0) + L_{MR}^k\right\}
\end{align}
\begin{align}
\label{eq:margin}
L_{MR}^k &=\sum_{l=k+1}\big\lbrace\mathds{1}_{[y_l=y_k]} \max(|a_k-a_l|-m^-,0)  
\nonumber \\
&+ \mathds{1}_{[y_l\neq y_k]} \max(m^+-|a_k-a_l|,0)\big\rbrace,    
\end{align}

where $m^+$ and $m^-$ are positive and negative margins, respectively. The above loss function consists of two parts: (i) The first part of the loss in eq.(\ref{eq:ce}) ensures that the proposals overlapping with the ground truth object locations are predicted as foreground with high confidence. (ii) The second part of the loss function, i.e., eq. (\ref{eq:margin}) is a margin-ranking loss that considers pairs of the proposals as input. It helps to further enforce a wider separation between foreground and background proposals in terms of the prediction probabilities, thereby improving the ranking of all the foreground proposals overlapping with the true location(s) of the object of interest. In addition to this loss, cross-entropy loss on the labeled (background or foreground) feature vectors of the region proposals and a regression loss on the predicted bounding box locations with respect to the ground truth bounding box are also used for training. 
\subsection{Multi-query Object Localization Setup}
\label{sec:multi}
The quality of the sketches can hinder the object localization performance. In many cases, sketches can be abstract and may not often contain the structural attributes that differentiate one object from another. The noise present in such sketches, along with their abstract nature, makes the task of sketch-guided object localization extremely challenging. However, we observe that the sketches are diverse in quality and when produced by different creators, they often tend to capture complementary information on an object's shape, attributes and appearance. This quality of complementarity can be leveraged towards improving the overall localization performance if information across these sketches is judiciously combined. With these motivations, we introduce the task of multi sketch query-guided object localization and present the following fusion strategies:\\
\noindent\textbf{Feature fusion:}
\label{sec:query_fus}
We observed complementarity across filter responses for different input sketches containing the same object. Such complementarity can be leveraged by a suitable feature fusion strategy to obtain less noisy and more holistic representation of the sketched object. In this paper, we used the global max pool operation to fuse feature maps of different sketch queries. Let $\{s_c^1,\dots, s_c^N\}$ be the set of $N$ sketches for the same object category $c\in C$ and  the representation learned by the backbone network for the $n^{th}$ sketch in this set is ${^n}s_c^{\phi_S}$. These feature maps are concatenated together to yield a composite feature map $R^{w\times h\times d\times N}$. A global max pool operation across all $N$ channels is subsequently performed to obtain a fused feature map for all the $N$ queries and the same is fed as input to the \emph{cross-modal attention} model.\\
\noindent\textbf{Attention Fusion:}
\label{sec:atten_fus}
In an alternative strategy, to attenuate noise in the attention maps produced by individual sketches, we concatenate these maps as generated by multiple queries and subsequently perform a depth-wise average pooling operation to obtain the final attention map. The fused attention map, thus generated, is used as input to the object localization pipeline (Section \ref{sec:scoring}).

\section{Experiments and Results}
\label{sec: expts}
\subsection{Datasets}
We use the following datasets for evaluating the performance of the proposed sketch-guided object localization framework.
\subsubsection{QuickDraw~\cite{jongejan2016quick}} is a large-scale sketch dataset containing 50 million sketches across $345$ categories. Sketches drawn from a subset of this dataset (please refer to the next paragraphs) are used as query. QuickDraw sketches are stored as vector graphics and are rasterized before feeding into ResNet.
\subsubsection{MS-COCO~\cite{lin2014microsoft}} is a large-scale image dataset which has been extensively used in object detection research. It has a total of $81$ classes, including background class, with dense object bounding box annotations. There are a total of $56$ classes which are common between MS-COCO and QuickDraw datasets. Therefore, in our object localization experiments, we randomly selected a total of $800K$ sketches across these common classes. We trained our model on COCO \textit{train2017} and evaluated on COCO \textit{val2017} dataset.
\subsubsection{PASCAL VOC~\cite{Everingham10}} is a popular object detection dataset with a total of $20$ classes. Among these, nine classes common with the QuickDraw dataset are selected for our experiments. Our model is trained on the union of VOC2007 \textit{train-val} and VOC2012 \textit{train-val} sets and evaluated on VOC \textit{test2007} set. 

\begin{table}[!t]
\renewcommand{\arraystretch}{1.2}
    \centering
    \begin{tabular}{lcc}
    \hline
    Model& mAP&  $\%$AP@50    \\ \hline
    Modified Faster R-CNN& 0.18&31.5\\
    Matchnet~\cite{hsieh2019one}& 0.28& 48.5\\
    Cross-modal attention (this work)& \textbf{0.30}&\textbf{50.0}\\
    \hline
    \end{tabular}%
    \caption{\label{tab:coco_full}
    \textbf{Results in one-shot common train-test categories setting on COCO \emph{val2017} dataset.} Comparison of various object localization baseline methods with the proposed \textit{cross-modal attention} model for sketch-guided object localization. We clearly outperform both the baselines based on mean average precision as well as $\%$ AP@50. For more details please refer to Section \ref{sec:res_full}.}
    \end{table}
\subsection{Baselines}
We adapt the following popular models from the object detection and image-guided localization communities as baselines towards evaluating sketch-guided object localization performance and compare them with our proposed method.
\subsubsection{Faster R-CNN~\cite{ren2015faster}} We adapt this object detector for query-guided object localization task. To this end, during training, we assign a $1$ (or $0$) class label to the object instance in the image if it belongs to the same (or different) class as the sketch query, and generate object proposals. Then, a binary classifier is used to classify each proposal as background or foreground. The query features are first concatenated with the region of interests(ROIs) features (pooled from region proposals) before passing it through the binary classifier. Additionally, we used a triplet loss to rank the object region proposal with respect to the sketch query. We refer to this baseline as \emph{modified Faster R-CNN}.
\subsubsection{Matchnet~\cite{hsieh2019one}} is a one-shot object localization method via image query. Here, non-local neural networks~\cite{wang2018non} and channel co-excitation~\cite{hu2018squeeze} were used to incorporate the query information in the image feature maps. We adapt this recent method to directly work with a sketch query and treat it as a second baseline.

For both these baseline methods, the feature extractors for sketches and images are ResNet-50 models pretrained on Imagenet and QuickDraw respectively.
     \begin{table}[!t]
     \renewcommand{\arraystretch}{1.2}
    \centering
    \begin{tabular}{lc}
    \hline
    Model& mAP         \\ \hline
    Modified Faster R-CNN& \textbf{0.65}\\
    Matchnet~\cite{hsieh2019one}& 0.61\\
    Cross-modal attention (this work)& \textbf{0.65}\\
    \hline
    \end{tabular}%
    \caption{\label{tab:voc_full}
    \textbf{Results in one-shot common train-test categories setting on VOC \emph{test2007} dataset.} Comparison of various object localization baseline methods with the proposed \textit{Cross-modal attention} model for sketch-guided object localization. We clearly outperform the state-of-the-art object localization method~\cite{hsieh2019one} and the results are comparable to Faster R-CNN based baseline. Please refer to Section~\ref{sec:osdtt} for more details.} 
    \end{table}   
\subsection{Experimental Setup}
\label{sec:setup}
We used two ResNet-50 models pre-trained on Imagenet~\cite{deng2009imagenet} and 5 million images of QuickDraw~\cite{jongejan2016quick} to get the feature representation for images and sketches respectively. The images from MS-COCO and PASCAL VOC datasets are used as target images, and the sketches randomly-drawn from the common classes of QuickDraw are used as queries. We evaluate performance of our model under the following four settings:
(i) one-shot common train-test categories, (ii) one-shot disjoint train-test categories, (iii) multi-query common train-test categories, and (iv) multi-query disjoint train-test categories.
\subsubsection{Disjoint train-test experimental setting:} Out of the 56 classes common across COCO and QuickDraw datasets, 42 and 14 classes are arbitrarily picked as `seen’ and `unseen’ categories, respectively. To ensure a one-shot disjoint train-test setting, the `seen’ and ‘unseen’ splits are mutually exclusive in terms of object classes as well as labeled bounding boxes present. Our model is trained solely on the ‘seen’ classes and only the `unseen’ classes are used for one-shot evaluation. Similarly for PASCAL VOC dataset, out of $9$ classes common with QuickDraw, 3 and 6 are arbitrarily picked as `unseen' and `seen' categories respectively. The image encoder is pre-trained using Imagenet dataset excluding 14 `unseen’ as well as all related classes to these 14 classes obtained by matching their WordNet synsets. Similarly, the sketch encoder is pre-trained using all the QuickDraw classes except the 14 categories in the unseen set.
\subsection{Results and Discussion}
In this section, we report results on the four experimental settings discussed in Sec.~\ref{sec:setup}. We then provide extensive discussions around merits and limitations of our proposed method in comparison to the chosen baselines. 
\begin{table}[!t]
\renewcommand{\arraystretch}{1.2}
        \centering
        \begin{tabular}{lcc}
        \hline
        Model&  Unseen Classes & Seen Classes         \\ \hline
        Modified Faster R-CNN& 7.4&34.5\\
        Matchnet ~\cite{hsieh2019one}& 12.4& \textbf{49.1}\\
        Cross-modal attention (this work)& \textbf{15.0}&48.8\\
        \hline
        \end{tabular}%
        \caption{\label{tab:oneshot_coco} 
       \textbf{Results in one-shot disjoint train-test categories setting on MS-COCO \textit{val2017} dataset.}  We report $\%$ AP@50 scores in this table. Here, \textit{unseen classes} contain the evaluation images and sketches of object categories that are not seen during training. Our proposed method significantly outperforms other baselines on unseen classes while achieving reasonably high performance for seen categories as well. Please refer to Section~\ref{sec:osdtt} for more details.}
        \end{table}
\newline
\newline
\noindent\textbf{One-shot common train-test categories:}
\label{sec:res_full}
The results for the proposed method in this setting are shown in Table~\ref{tab:coco_full} for MS-COCO dataset. As shown, the proposed method outperforms both the baselines significantly. 
When compared to the modified Faster R-CNN baseline, our method performs significantly better. This is primarily because, unlike Faster R-CNN, our \textit{cross-modal attention} framework effectively incorporates information from the query using spatial compatibility (attention) map to generate relevant region proposals. Moreover, our method also outperforms the Matchnet baseline because the non-local feature maps and channel co-excitation module in Matchnet are sensitive towards the domain gap present between query and image feature maps in our task. On the contrary, our \emph{cross-modal attention} framework addresses this by explicitly computing a spatial compatibility map (attention map).

\begin{figure}[!t]
    \centering
    \includegraphics[width = 12 cm]{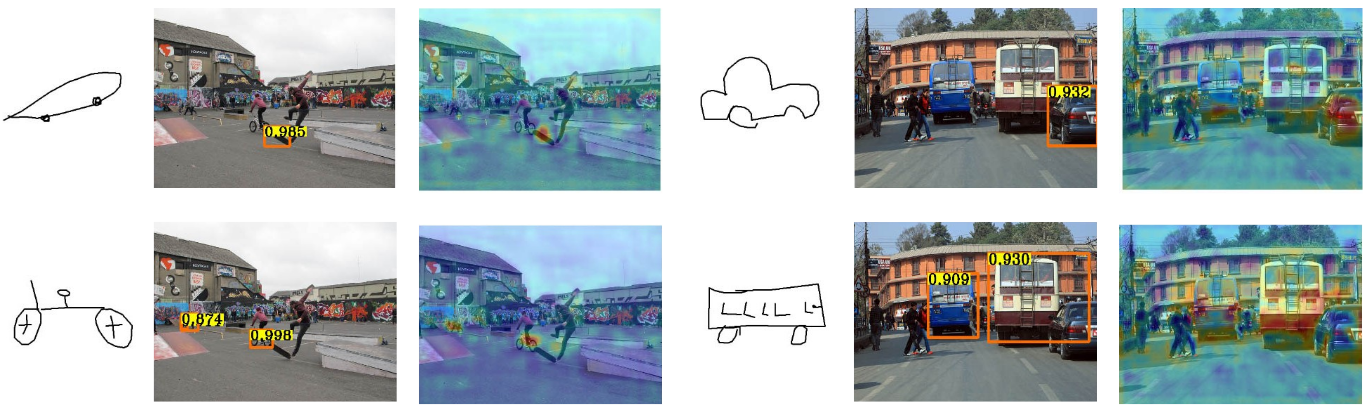}
    \caption{\label{fig:attention} Sketch queries and object localization results are shown along with the corresponding attention maps. The attention map produces high compatibility scores (shown in red color) for the regions on the natural image that contains object same as that in the sketch query. This leads to the localization (shown using a red bounding box on input image) of the relevant object. Please note that sketches have been enhanced for better visualization. \textbf{[Best viewed in color]}.}
\end{figure}

The results on the PASCAL VOC dataset are reported in Table~\ref{tab:voc_full}. PASCAL VOC contains less number of training images ($\approx$ 9K) with small variability across training classes in our setting (only 9 classes in the training set). The Faster R-CNN baseline is comparable to the proposed method indicating that sketch query-guided object localization is challenging without sufficient data. The Matchnet baseline model degrades in performance, indicating its inability to incorporate sketch information during proposal generation in the case of small data size and large domain gap.
\newline
\newline
\noindent\textbf{One-shot disjoint train-test categories:}
\label{sec:osdtt}
In this setting, models are evaluated on the object categories which are not seen during training. The results are provided in Table~\ref{tab:oneshot_coco} for MS-COCO dataset. We have selected model checkpoints that perform the best on \emph{unseen} categories for each model. It is evident that one-shot object localization under disjoint train-test object categories is a hard problem as it leads to degradation in performance for all three models. Faster R-CNN baseline suffers most degradation because it does not incorporate sketch information during proposal generation and perform poorly in proposal generation for unseen categories. While Matchnet outperforms the modified Faster R-CNN baseline, it still performs significantly worse than the proposed method.
Contrary to baseline methods, our model, by virtue of \emph{cross-modal attention}, embeds query information in the image feature representation before generating region proposals. Therefore, it is inherently capable of generating relevant object proposals even when attempting to localize unseen object classes. As a result, our method reports superior performance on unseen categories while achieving reasonably high performance for seen categories as well. 
\begin{table}[!t]
\renewcommand{\arraystretch}{1.2}
            \centering
            \begin{tabular}{lcccc}
            \hline
            Our models& Unseen (D) & Seen (D) & All (C) & mAP (C)       \\ \hline
            Cross-modal attention& 15.0& 48.8&50.0&0.30\\
             $\,\,\,$  +Query Fusion(5Q) & 16.3&\textbf{52.2} &52.6&\textbf{0.32}\\
             $\,\,\,$  +Attention Fusion(5Q) & \textbf{17.1}& 51.9&\textbf{53.1}&\textbf{0.32}\\
            \hline
            \end{tabular}%
            \caption{\label{tab:few_shot_coco}
            \textbf{Results in multi-query common and disjoint train-test categories setting on \textbf{MS-COCO \emph{val2017} dataset.} }Comparison of the proposed fusion strategies for $5$ sketch queries. Here, $5Q$, (C) and (D) represents $5$ sketch queries, common train and test categories, and disjoint train and test categories respectively. $\%$ AP@50 is reported except the last column which reports mAP.}
\end{table}
\newline
\newline
\noindent\textbf{Multi-query common and disjoint train-test categories:}
\label{sec:res_multi}
Here we discuss the effect of utilizing multiple sketch queries on the object localization performance. We evaluated our model using two fusion strategies presented in section \ref{sec:multi} under the following two settings: (a) common train and test categories, (b) completely disjoint train and test object classes. The evaluation results of these fusion strategies are provided in Table~\ref{tab:few_shot_coco}. Both these fusion strategies provide consistent improvement in performance over the proposed cross-modal attention based baseline model, which just uses a single sketch query as input. This suggests that the fusion strategies are able to capture the complementary information present across the multiple sketches of the same object. Moreover, we also evaluated the proposed fusion strategies for sketch-guided object localization on \emph{unseen} categories and observe that even in such cases, these fusion strategies are able to aid the proposed localization framework.\\
\newline
\noindent\textbf{Utility of margin-based ranking loss:} We analyze the effect of different components of loss function defined in eq.(\ref{eq:ce}) and eq. (\ref{eq:margin}) on the model's performance using MS-COCO \textit{val2017} dataset. The model which is trained with only foreground-background classification loss, first part of eq.(\ref{eq:ce}) gives $\%AP@50$ of $40.0$\% as compared to $41.4$\%, achieved by the model trained on only the margin-ranking loss ($L_{MR}$) in eq.(\ref{eq:margin}). However, training the model on both these losses significantly improves the performance ($50.0~\%AP@50$).
\newline
\newline
\noindent\textbf{Visualizing the attention map and localization:}
In order to visualize the effect of sketch queries on the generated attention maps and the resulting localizations, we query an image with sketches of different object categories. As shown in Fig.~\ref{fig:attention}, our novel \textit{cross-modal attention} scheme is able to assign high compatibility scores to the ground-truth object locations in the natural image while at the same time ignoring instances of other object categories, thereby producing precise localizations. 
\section{Conclusion}
\label{sec: conc}
In this paper,  we have introduced a novel problem of sketch-guided object localization in natural images and presented an end-to-end trainable framework using novel \textit{cross-modal attention} for this task. The effectiveness of the proposed attention scheme is substantiated by the significant performance improvement achieved over the baseline methods on two public benchmarks. We have extended the framework to a multi-query setup and proposed two fusion strategies towards the same. The superior performance of our framework over the baseline in this setting corroborates that the proposed fusion strategies are able to leverage the complementarity present across multiple sketches of the same object. 
We look forward to exciting future research in sketch-guided computer vision tasks inspired by the problem that we have introduced.
\section*{Acknowledgements}
The authors would like to thank the Advanced Data Management Research Group, Corporate Technologies, Siemens Technology and Services Pvt. Ltd., and Pratiksha Trust, Bengaluru, India for partly supporting this research.

\bibliographystyle{splncs04}
\bibliography{egbib}
\newpage
\appendix
\section{Implementation Details}
We implemented proposed cross-modal attention model using the Pytorch v1.0.1 framework with CUDA 10.0 and CUDNN v7.1. The model is trained with stochastic gradient descent (SGD) with momentum of $0.9$ on one NVIDIA 1080-Ti with a batch-size of $10$. The learning rate was initially set at $0.01$ but it decays with a rate of $0.1$ after every four epochs, and it is trained for $30$ epochs. The constant $\cal K$ in Equation 4 is fixed at $256$ and $m^+=0.3$ and $m^-=0.7$ in Equation 7 and 8 for all experiments.

For optimal results, \emph{cross-modal attention} model is trained incrementally. Firstly, the localization model is trained without attention. Then, the attention model is added to it, and it is trained again. The training protocol is same as explained before and it is same for both the steps.


\section{Additional Results}
\begin{figure}[!h]
    \centering
    \includegraphics[width = 10cm]{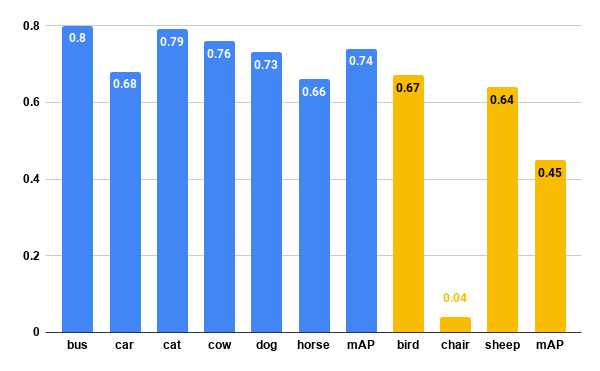}
    \caption{\label{fig:query_fusion} \textbf{Class-wise sketch-guided object localization results of our model on VOC \emph{test2007} dataset are shown in this plot.} In this experiment, training and testing classes are disjoint. Bar-plot in `blue' color represents results on seen categories and the plots in `yellow' represents unseen categories. The class-wise AP values are reported and the mAP is also reported separately for seen and unseen categories. \textbf{[Best viewed in color]}}
    \label{fig:bar_oneshot}
\end{figure}

\begin{figure}[!h]
    \centering
    \includegraphics[width = 10cm]{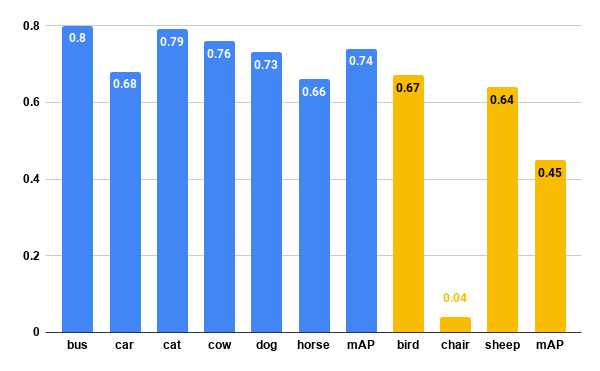}
    \caption{\label{fig:query_fusion} \textbf{Class-wise sketch-guided object localization results of our model on VOC \emph{test2007} dataset are shown in this plot.} The class-wise AP values are reported along with the mAP. \textbf{[Best viewed in color]}.}
    \label{fig:bar_fullshot}
\end{figure}

In this section, we describe additional results on single query (one-shot) sketch-guided localization as well as multi-query sketch-guided localization with $3$ sketch queries on Pascal VOC dataset. The results are reported for seen and unseen sets of disjoint ($C_{train}\cap C_{test}=\phi$) training and test sets as well as the union of both the sets. The results are shown in Fig~\ref{fig:bar_oneshot} and \ref{fig:bar_fullshot}. 

For the disjoint train-test classes experiment, the data from $6$ classes is used for training and the data from $3$ classes is used for evaluation. There is no overlap between training and testing images.

        

The multi-query sketch-guided localization model has also been evaluated for the case of $3$ sketch queries. We evaluated the model for the disjoint train and test classes as well as the common train and test classes. Both the \emph{Query Fusion} and \emph{Attention Fusion} methods show consistent improvement across disjoint classes and common classes. The results are reported in table \ref{tab:few_shot_coco_3}. It is evident that both proposed fusion methods are able to effectively combine complementary information present in multiple sketches with as little as $3$ sketch queries.  
\begin{table}[!h]
            \centering
            \begin{tabular}{lcccc}
             \toprule
            Our models& Unseen (D)~~~~~~ & Seen (D)~~~~~~ & All (C)~~~~~~ & mAP (C)       \\ 
            \midrule
            Cross-modal attention& 15.0& 48.8&50.0&0.30\\
             $\,\,\,\,\,\,$  +Query Fusion(3Q) & 17.1&\textbf{51.4} &51.9&\textbf{0.31}\\
             $\,\,\,\,\,\,$  +Attention Fusion(3Q) & \textbf{17.6}& 50.9&\textbf{52.0}&\textbf{0.31}\\
            \bottomrule
            \end{tabular}%
            \caption{\label{tab:few_shot_coco_3}
            \textbf{Results in multi-query common and disjoint train-test categories setting on \textbf{MS-COCO \emph{Val2017} dataset.} }Comparison of the proposed fusion strategies for $3$ sketch queries. Here, $3Q$, (C) and (D) represents $3$ sketch queries, common train and test categories, and disjoint train and test categories respectively. $\%$ AP@50 is reported except where mAP is mentioned. For more details refer Section 4.4 and 3.4.}
            \end{table}

\section{Additional Visualizations}

\begin{figure}[!h]
    \centering
    \includegraphics[width = 12cm]{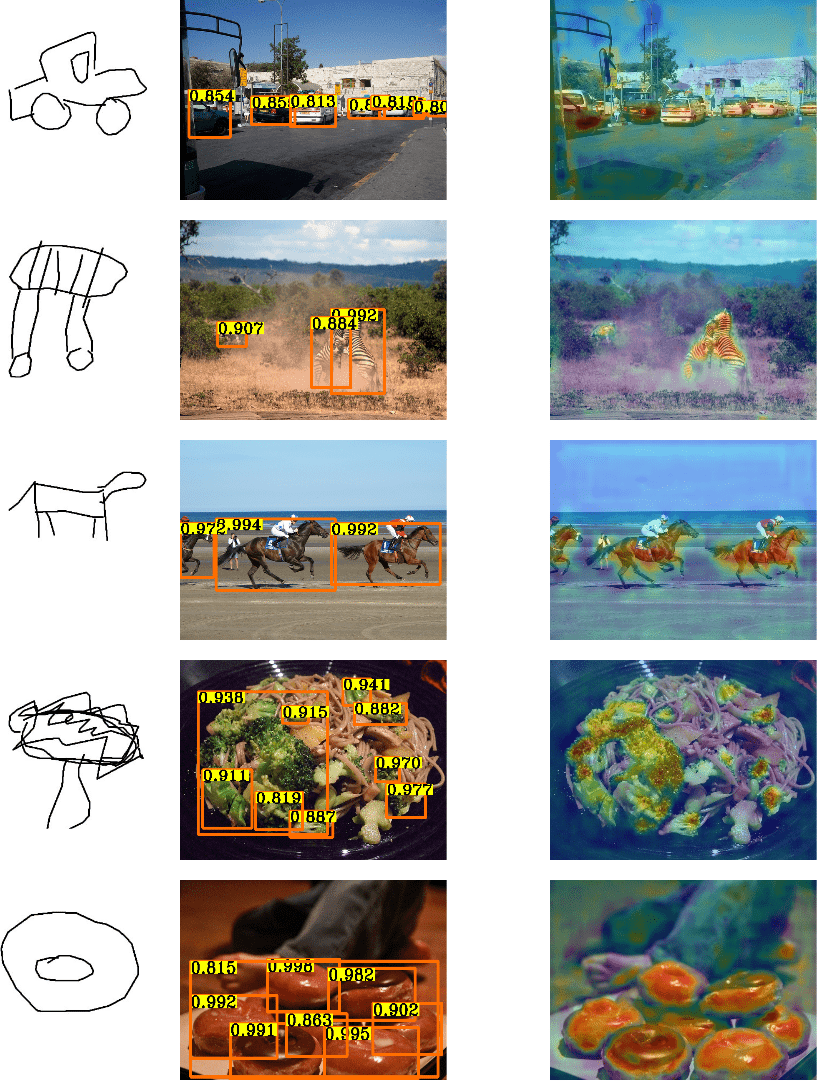}
    \caption{\label{fig:query_fusion} Sketch query and object localization are shown along with the corresponding attention maps. The images which have multiple instances of the same class as query is shown along with the corresponding attention maps. The image on the left is a sketch query, the image in the center contains the localization generated and the image on the right is the corresponding attention map. As we can see, \emph{cross-modal} attention is able to focus on multiple instance of the same class. \textbf{[Best viewed in color]}.}
\end{figure}

\begin{figure}[!h]
    \centering
    \includegraphics[width=12cm]{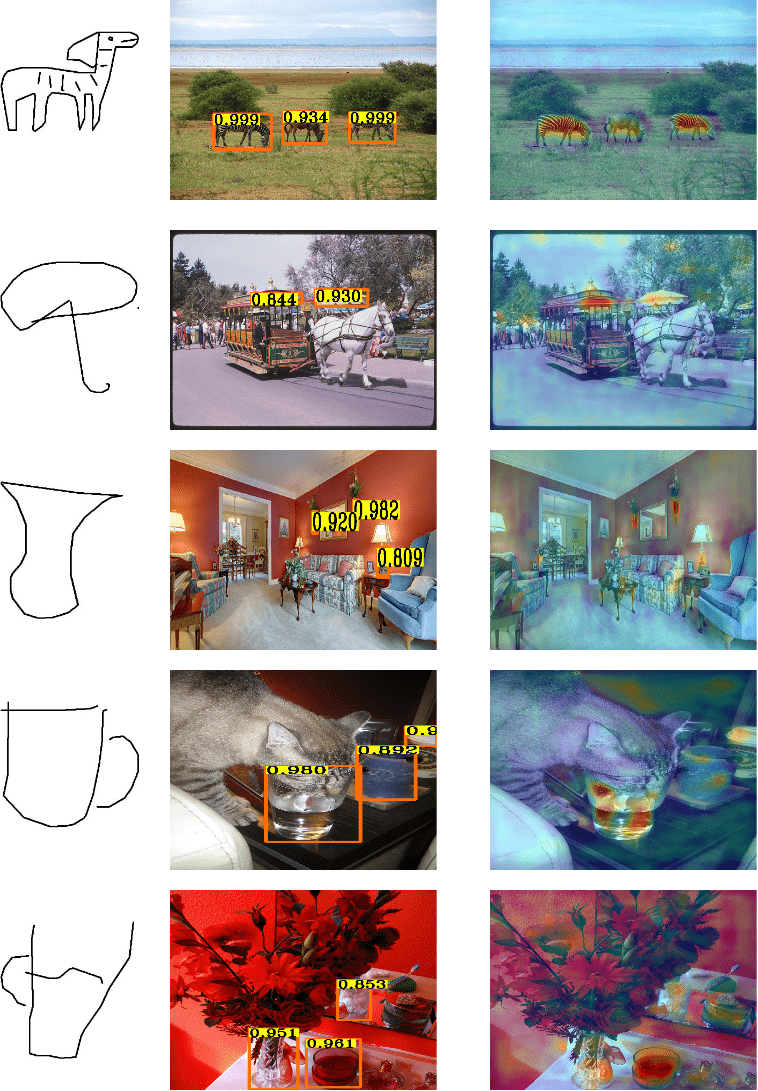}
    \caption{\label{fig:query_fusion} A few failure cases are illustrated here. The image on the left is a query sketch, the image in the center contains the generated localizations and the image on the right is the corresponding attention map. The model localized objects of wrong classes but upon closer look, the false detection look similar to the objects of the correct class. The query sketches in this case are not visually rich enough the distinguish the correct and the false detection. \textbf{[Best viewed in color]}.}
\end{figure}
\end{document}